%% file: sample-acmtog-SIGGRAPH-submission.tex
\documentclass[sigconf]{acmart}
\acmSubmissionID{1970}

\usepackage{booktabs} 
\usepackage{amsmath}
\usepackage{multirow}
\usepackage{graphicx}
\usepackage{nicefrac}
\usepackage{enumitem}
\usepackage{hyperref}
\usepackage{soul} 
\usepackage{xcolor} 
\usepackage[table]{xcolor}
\usepackage{caption}

\sethlcolor{yellow} 
\usepackage[most]{tcolorbox}
\AtBeginDocument{%
  }
\usepackage[ruled]{algorithm2e} 

\SetAlFnt{\small}
\SetAlCapFnt{\small}
\SetAlCapNameFnt{\small}
\SetAlCapHSkip{0pt}

\acmJournal{TOG}

\copyrightyear{2025} 
\acmYear{2025} 
\setcopyright{acmlicensed}
\acmConference[SA Conference Papers '25]{SIGGRAPH Asia
 2025 Conference Papers}{December 15--18, 2025}{Hong Kong, Hong Kong}
\acmBooktitle{SIGGRAPH Asia 2025 Conference Papers (SA Conference Papers '25),
December 15--18, 2025, Hong Kong, Hong Kong}
\acmDOI{10.1145/3757377.3763954}
\acmISBN{979-8-4007-2137-3/2025/12}

\begin{document}
\title{Uni-Inter: Unifying 3D Human Motion Synthesis Across Diverse Interaction Contexts}
\author{Sheng Liu}
\authornote{Both authors contributed equally to this research.}
\affiliation{%
 \institution{Nanjing University}
 \city{Nanjing}
 \postcode{210023}
 \country{China}}
\email{anderliu@smail.nju.edu.cn}
\author{Yuanzhi Liang}
\authornotemark[1]
\affiliation{%
 \institution{Institute of Artificial Intelligence, China Telecom (TeleAI)}
 \city{Shanghai}
 \country{China}
}
\email{liangyzh18@outlook.com}

\author{Jiepeng Wang}
\affiliation{%
\institution{Institute of Artificial Intelligence, China Telecom (TeleAI)}
\city{Shanghai}
\country{China}}
\email{jiepeng@connect.hku.hk}
\author{Sidan Du}
\authornote{Corresponding author}
\affiliation{%
 \institution{Nanjing University}
 \city{Nanjing}
 \country{China}
}
\email{coff128@nju.edu.cn}
\author{Chi Zhang}
\authornotemark[2]
\affiliation{%
 \institution{Institute of Artificial Intelligence, China Telecom (TeleAI)}
 \city{Shanghai}
 \country{China}}
\email{zhangc120@chinatelecom.cn}
\author{Xuelong Li}
\authornotemark[2]
\affiliation{%
 \institution{Institute of Artificial Intelligence, China Telecom (TeleAI)}
 \city{Shanghai}
 \country{China}
}
\email{xuelong_li@ieee.org }


\begin{abstract}
We present Uni-Inter, a unified framework for human motion generation that supports a wide range of interaction scenarios: including human-human, human-object, and human-scene—within a single, task-agnostic architecture. In contrast to existing methods that rely on task-specific designs and exhibit limited generalization, Uni-Inter introduces the Unified Interactive Volume (UIV), a volumetric representation that encodes heterogeneous interactive entities into a shared spatial field. This enables consistent relational reasoning and compound interaction modeling. Motion generation is formulated as joint-wise probabilistic prediction over the UIV, allowing the model to capture fine-grained spatial dependencies and produce coherent, context-aware behaviors. Experiments across three representative interaction tasks demonstrate that Uni-Inter achieves competitive performance and generalizes well to novel combinations of entities. These results suggest that unified modeling of compound interactions offers a promising direction for scalable motion synthesis in complex environments. Our code can be found at \url{https://github.com/Darkdawner/Uni-Inter}.
\end{abstract}

%
%

\ccsdesc[300]{Computing methodologies~Motion processing}
\ccsdesc[300]{Computing methodologies~Motion capture}
\ccsdesc[300]{Computing methodologies~Computer vision problems}
%
%

\keywords{Unified Motion Generation, Human-Object Interaction, Human-Scene Interaction, Human-Human Interaction}

\begin{teaserfigure}
  \centering
  \includegraphics[width=\textwidth]{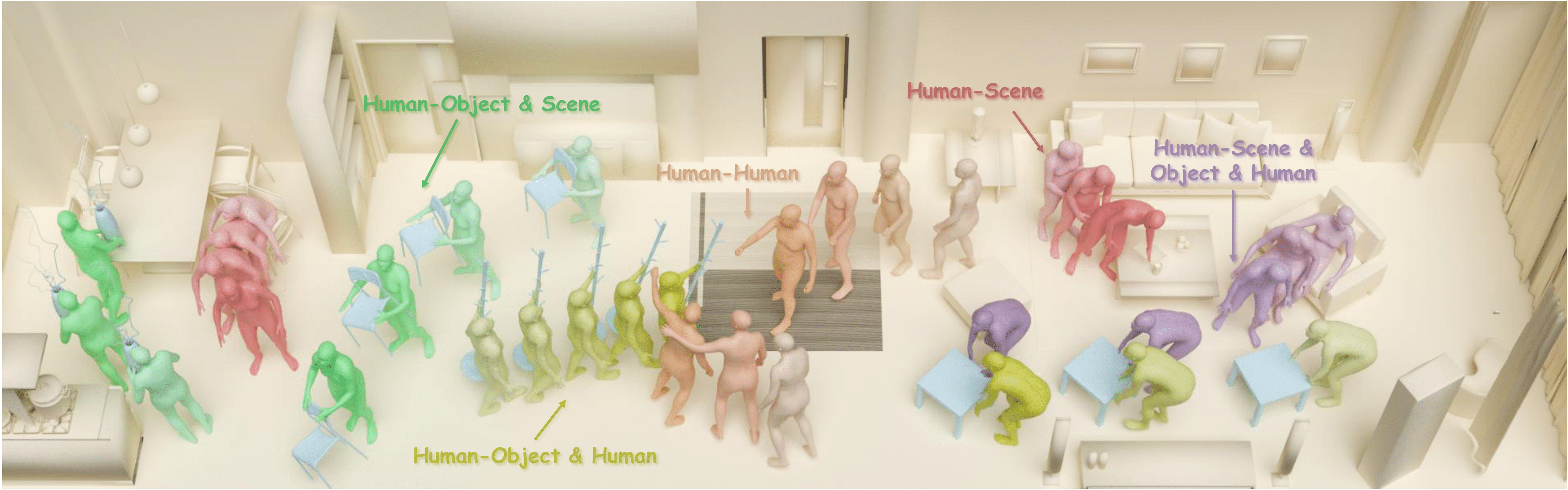}
  \caption{Uni-Inter supports interaction motion generation with arbitrary combinations of humans, objects, and scene elements. Leveraging a unified interactive volume and a task-agnostic modeling paradigm, Uni-Inter generalizes across diverse interaction types, producing coherent and context-aware motions in compound real-world scenarios. Best viewed in color and with zoom.}
  \label{fig:teaser}
\end{teaserfigure}

\maketitle

\input{samplebody-journals}

\end{document}

%% file: samplebody-journals.tex
\section{Introduction}
\setlength{\textfloatsep}{5pt}
\setlength{\intextsep}{5pt} 
\begin{figure*}[!t]
  \centering
  \includegraphics[width=\linewidth]{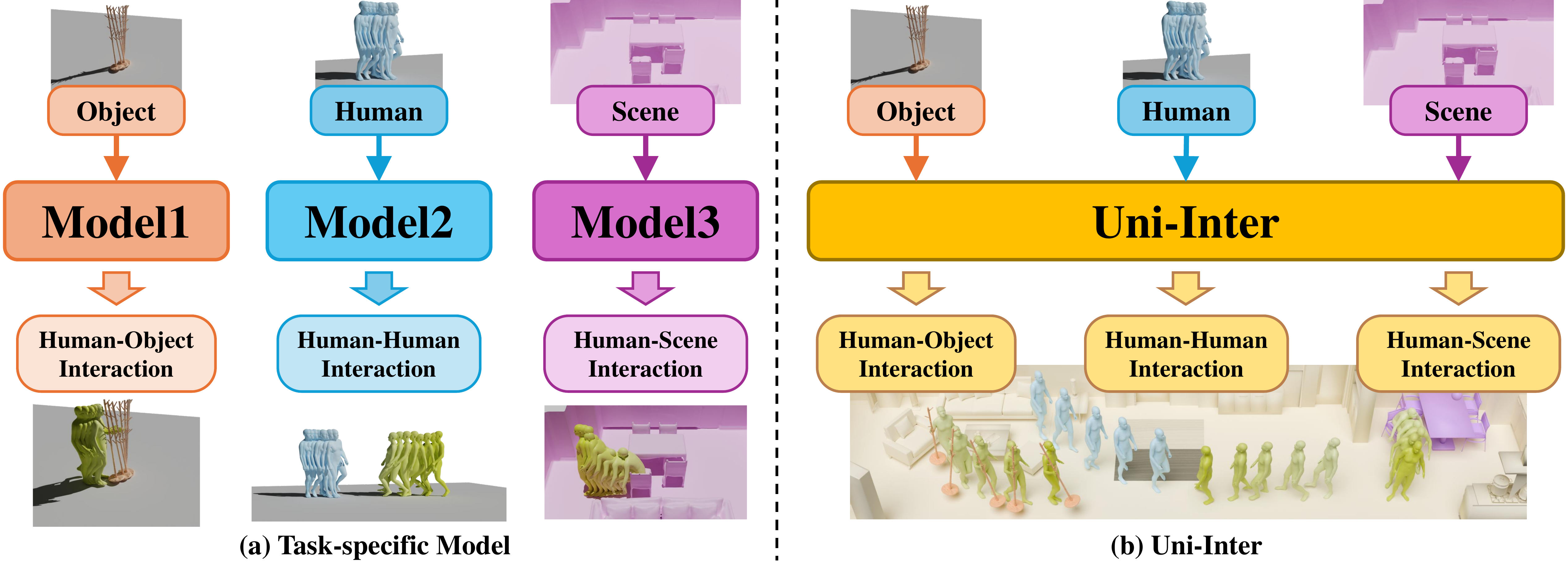}
  \captionsetup{skip=2pt} 
  \caption{Different paradigms for compound interaction motion generation. (a) Existing methods rely on task-specific architectures, resulting in \textbf{separately} modeling when handling compound interactions involving multiple entity types. (b) In contrast, Uni-Inter provides a \textbf{unified} motion generation framework that seamlessly supports arbitrary combinations of interactive entities—including humans, objects, and scenes—within a single model.}
  \label{fig:intro}
\end{figure*}

Modeling human interactions in dynamic environments has emerged as a key challenge at the intersection of computer graphics, vision, and embodied AI. Interaction motion generation—the task of synthesizing contextually appropriate human motion in response to surrounding entities—plays a pivotal role in applications such as character animation, immersive virtual environments, and assistive robotics. Effective systems must interpret scene semantics~\cite{li2023voxformer, wei2023surroundocc, jiang2024symphonize, wang2022towards, jiang2024scaling}, coordinate across multiple agents and objects~\cite{liang2024intergen, ng2023diffusion, li2024controllable}, and generate plausible motion trajectories~\cite{tevet2022human, wang2017optimal, xu2024matrix} that are both physically valid and semantically aligned.

Recent advances have led to strong performance on specific subproblems, such as human-human \cite{liang2024intergen, shafir2023human, starke2020local, starke2021neural, xu2023actformer}, human-object \cite{jiang2023full, kulkarni2024nifty, pi2023hierarchical, zhang2022couch, li2023object, li2024controllable}, and human-scene interaction \cite{araujo2023circle, pan2024synthesizing, wang2022towards, jiang2024scaling}. However, these tasks are typically treated in isolation. In contrast, real-world scenarios often involve compound interactions—simultaneous engagement with multiple entity types. For instance, a person might pick up two cups (human-object), hand one to a friend (human-human), while simultaneously sitting down on a sofa (human-scene), all within a continuous motion sequence. Such interactions inherently involve overlapping physical constraints~\cite{liu2024revisit, han2025reindiffuse} (e.g., collision, support), social dynamics~\cite{capozzi2018attention,parkinson2008emotions, rutten2017beyond} (e.g., proximity, attention), and task semantics~\cite{park2004semantic, kuehne2014language, pustejovsky2022multimodal} (e.g., handover, sitting).

This raises a critical limitation in existing methods: their inability to generalize to such compound interaction scenarios. In adjacent domains like autonomous driving, unified perception frameworks \cite{li2023voxformer, wei2023surroundocc, jiang2024symphonize} have demonstrated the value of modeling pedestrians, vehicles, and road structures within a shared representational space. This motivates the question: Can interaction motion generation benefit from a similarly unified modeling approach to better handle compound real-world behaviors?

One core challenge is the lack of a shared representation for heterogeneous entities. Existing systems use disparate formats (e.g., skeletons for humans, point clouds for objects, occupancy grids for scenes), making it difficult to jointly capture the physical constraints, social dynamics, and task semantics that naturally co-occur in compound interactions. This disconnect often results in incoherent motion outputs when multiple interaction types are involved. Moreover, most prior work \cite{liang2024intergen, shafir2023human, li2024controllable, wang2022towards, jiang2024scaling} adopts task-specific training paradigms, where models are optimized for narrowly defined interaction types using separate pipelines. These specialized designs limit cross-task generalization and hinder scalability to complex multi-agent environments.

In this paper, we introduce Uni-Inter, a unified multi-task framework for compound interaction motion generation. At its core is the Unified Interactive Volume (UIV), a volumetric representation that maps humans, objects, and scene elements into a shared 3D occupancy field. Inspired by semantic occupancy grids used in scene understanding, UIV serves as a spatial-semantic substrate that enables coherent modeling of diverse interaction types under a common structure. UIV allows the system to reason jointly over physical constraints, social dynamics, and task semantics in a unified format.

To fully leverage this representation, we further design a motion modeling strategy aligned with UIV. Rather than predicting joint coordinates or rotations independently, Uni-Inter models joint-wise spatial distributions over the interaction volume, enabling context-aware trajectory generation that respects the surrounding geometry and interaction context. This formulation transforms motion generation into a structured spatial inference problem, leading to stronger generalization across tasks and more coherent behavior in complex multi-entity scenes. 

Our main contributions are as follows:

1. We introduce \textbf{Uni-Inter}, a unified framework for interaction motion generation that supports arbitrary combinations of humans, objects, and scenes within a single cohesive system. To our knowledge, this is the first approach to model compound interactions under a shared representation.

2. A novel representation, the \textbf{Unified Interactive Volume (UIV)}, is introduced to encode heterogeneous interactive entities into a shared voxel-based space. This spatial-semantic abstraction facilitates consistent relational reasoning across interaction types.

3. We propose \textbf{UIV-aligned Regularization}, formulating motion prediction as spatial distribution estimation over the interaction volume. This supports structured inference and improves coherence in complex scenarios.

4. Extensive experiments are conducted across multiple benchmarks, demonstrating that Uni-Inter outperforms prior methods in standard interaction settings, with strong generalization to unseen combinations of entities.

\section{Related Work}
\noindent \textbf{Human-Object Interaction.} Several recent methods \cite{jiang2023full, kulkarni2024nifty, pi2023hierarchical, zhang2022couch} address human-object interaction by generating plausible motion sequences. Nevertheless, they often limit their scope to static objects, which constrains the diversity of possible interactions. Alternative approaches generate human interaction motions by conditioning on textual descriptions \cite{peng2025hoi, wu2024thor, xu2024interdreamer, wu2024human} or predefined contact points \cite{diller2024cg}. OMOMO \cite{li2023object} proposes a conditional diffusion model that generates full-body motions guided by object trajectories, yet it primarily targets scenarios involving small-scale hand-object interactions. IMoS \cite{ghosh2023imos} extends this line of work by synthesizing both human and object motion from textual inputs, but similarly focuses on fine-grained manipulations involving small items. InterDiff \cite{xu2023interdiff} considers full-body interactions with moving objects, but formulates the task as future motion prediction based on prior trajectories and object information. CHOIS \cite{li2024controllable} attempts to guide motion synthesis using both object movement and language text. However, the realism and quality of the generated interactions still leave room for improvement.

\noindent \textbf{Human-Human Interaction.} Modeling human-human interactions has garnered growing interest in recent years \cite{liang2024intergen, shafir2023human, starke2020local, starke2021neural, xu2023actformer}. Some approaches treat both agents symmetrically, without accounting for role asymmetry \cite{liang2024intergen, xu2023actformer}, while others limit their scope to specific action categories typically used in graphics applications \cite{starke2021neural}. ReGenNet \cite{xu2024regennet} leverages a conditional diffusion framework to generate responsive motions of one person based on the other’s actions. However, its design is constrained to dyadic interaction and cannot generalize to broader multi-person interaction scenarios.

\noindent \textbf{Human-Scene Interaction.} Human-scene interaction generation spans a range of modeling targets, from static pose estimation \cite{li2019putting, zhang2020generating1, zhang2020generating2} to dynamic motion synthesis over time \cite{araujo2023circle, huang2023diffusion,mir2024generating, pan2024synthesizing, wang2022towards}. Generative frameworks such as conditional VAEs \cite{sohn2015learning} and diffusion models \cite{ho2020denoising, song2020denoising} have been widely adopted to produce diverse and realistic motions under specified constraints. Depending on how motion is modeled, some methods directly sample full sequences holistically, while others adopt autoregressive decoding \cite{cao2020long, corona2020context, hassan2021stochastic} or anchor-based interpolation schemes \cite{wang2021synthesizing} for scalability and temporal flexibility. Trumans \cite{jiang2024scaling} adopts a diffusion-based approach to generate short motion fragments, which are then autoregressively composed during inference to enable the synthesis of high-quality motions of arbitrary lengths. In parallel, controllability has been enhanced through the integration of high-level semantics, such as structured action categories \cite{zhao2022compositional} or natural language guidance \cite{wang2022humanise, xiao2023unified, yi2024generating}, enabling models to align generated motions with human intent or task semantics. Recent methods \cite{liu2024revisit, lu2024choice} have further extended the scope to model human–environment interactions, encompassing both object- and scene-level contexts.
\section{Preliminary}
\label{diffusion}
\begin{figure*}[!t]
  \centering
  \includegraphics[width=\linewidth]{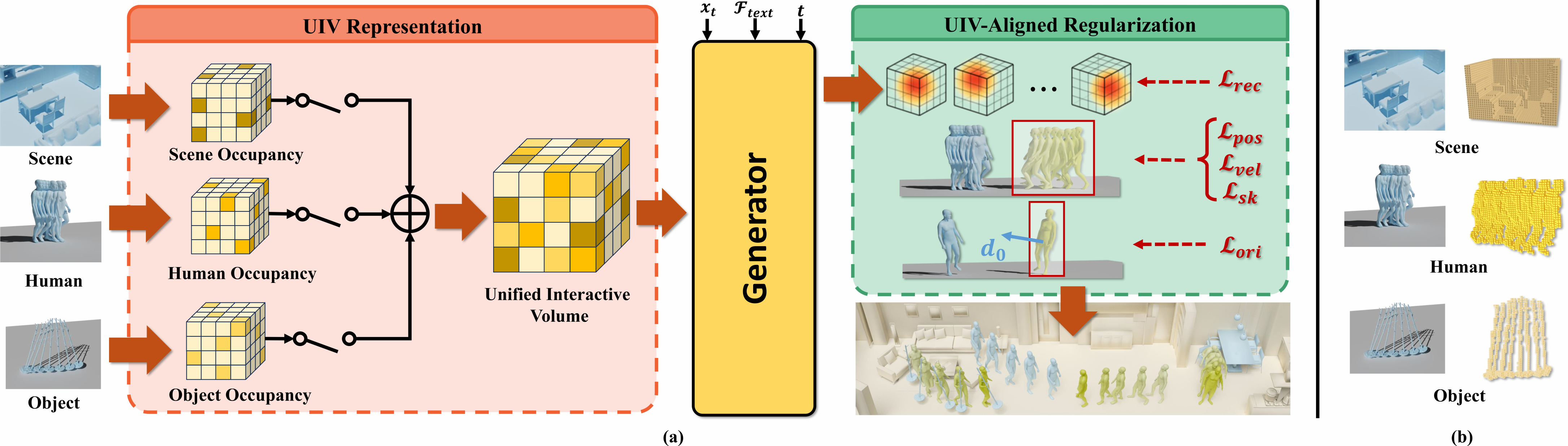}
  \captionsetup{skip=2pt} 
  \caption{(a) Uni-Inter supports arbitrary combinations of interactive entities as input and generates corresponding interaction motions. This is enabled by the Unified Interactive Volume (UIV) representation and UIV-aligned regularization. Each interaction entity—whether human, object, or scene—is first encoded as a semantic occupancy grid in the interaction space and then merged into the UIV, which serves as the conditional input to the motion generator. The generator predicts joint-wise spatial distributions guided by the carefully designed UIV-aligned regularization, enabling coherent and context-aware motion synthesis. (b) Illustration of voxel-based representations for different interaction entities, including humans, objects, and the surrounding scene.}
  \label{fig:method}
\end{figure*}

Recent advances in diffusion models have shown strong capability in synthesizing high-fidelity human motion through a probabilistic denoising process. Inspired by their success in image and text-to-image synthesis \cite{ramesh2021zero, rombach2022high, saharia2022photorealistic}, diffusion-based motion generation methods \cite{tevet2022human, zhang2023remodiffuse, zhang2024motiondiffuse} adopt a non-autoregressive paradigm, enabling parallel generation of complete motion sequences while naturally ensuring long-term temporal consistency and physical plausibility.

Diffusion modeling learns to invert a fixed stochastic process: Gaussian noise is progressively added to a ground-truth motion sequence $x_0^*$ until it becomes nearly pure noise $x_N$. The model then learns to iteratively denoise $x_N$ back to $x_0$, parameterizing each reverse step as:
\begin{equation}
    P_\theta(x_{i-1}|x_i)=\mathcal{N}(\mu_i(\theta), \sigma_i^2I)
\end{equation}
where $\mu_i(\theta)$ combines the estimated clean motion and the noisy input under a noise schedule $\alpha_i$. We directly regress to the clean motion $x_0$ using the reconstruction loss:
\begin{equation}
    \mathcal{L}_{rec}(x_0,x_0^*)=\Vert x_0-x_0^*\Vert_2^2
\end{equation}
At inference, motion is generated by sampling $x_N \sim \mathcal{N}(0,I)$ and iteratively applying the denoising process until $x_0$ is obtained.

\section{Unified Interaction Generation}
Our goal is to develop a unified model for motion generation that supports a wide range of tasks with minimal task-specific tuning. This shared architecture allows for joint training across diverse tasks and promotes effective knowledge transfer. Given a textual description $c_{text}$ and a conditioning observation $c$, our goal is to generate a motion sequence $m_{1:T}=\{j^{1:K}_1, j^{1:K}_2, \cdots, j^{1:K}_T\}$ that aligns with both the semantics of the text and the observed context. Formally, the motion is generated as $m_{1:T}=G(\epsilon;c_{text},c)$, where $\epsilon\sim \mathcal{N}(0,I)$ is a noise vector randomly sampled from a standard Gaussian distribution.

\subsection{Unified Interactive Volume}
\label{sec:unified}
Interactive motion generation consists of three core tasks: Human-Human, Human-Object, and Human-Scene interactions. To support a unified framework, it is necessary to represent all interactive entities-humans, objects, and scenes-within a shared, task-agnostic representations. In the field of autonomous driving, similar perceptual unification has been achieved using grid-based occupancy representations. Inspired by this, we extend this representation to the domain of motion generation, enabling a unified representation across diverse interacting entities with \textbf{U}nified \textbf{I}nteractive \textbf{V}olume (UIV). Let $\mathcal{S} \in \mathbb{R}^{H \times W \times D}$ denote a discretized 3D interaction space, where $H$, $W$, and $D$ are the spatial resolution along each axis. We assume that all interactions occur within this bounded volume, which is uniformly divided into voxels indexed by spatial coordinate $u$. That means UIV is a static, predefined space where the generated human moves. At each time step $t \in {1, \ldots, T}$, we define a semantic occupancy map ${\phi}_t : \mathbb{R}^3 \rightarrow \{0,1\}^3$, assigning each voxel either to a specific semantic entity type $c \in \mathcal{C}$ or marking it as empty:
\begin{equation}
\label{equ:1}
    \mathcal{\phi}_t({u}) = \sum_{c \in \mathcal{C}} \mathcal{I}_{c}(u, t)\cdot c
\end{equation}
where $\mathcal{C} \subset \mathbb{R}^3$ denotes the space of one-hot entity labels (e.g., scene, object, human). $\mathcal{I}_{c}$ is an indicator function defined as:
\begin{equation}
    \mathcal{I}_{c}(u, t) = \left\{\begin{array}{ll}
    1, & \text{voxel $u$ at time $t$ is occupied by entity of type $c$} \\
    0, & \text{otherwise}
    \end{array}\right.
\end{equation}
The function $\phi$ is applied to each voxel in $\mathcal{S}$, producing a semantically encoded volumetric representation $\mathcal{V}\in\mathbb{R}^{H\times W\times D\times 3}$. This representation aggregates all interactive entities over time into a unified interactive volume:
\begin{equation}
    \Omega = \{\mathcal{V}_t\}_{t=1}^{T}
\end{equation}

\noindent \textbf{UIV for Human.} Leveraging the SMPL model \cite{loper2015smpl}, we convert conditional human motions into dense mesh representations. We discretize the vertex positions in the space $\mathcal{S}$ and transform them into the corresponding volumetric representation as in Equation.~\ref{equ:1}, where $c_{h}=[1,0,0]$ for the human-type entity.

\noindent \textbf{UIV for Object \& Scene.} For object-type entities, we first compute the position of each frame's object vertices in $\mathcal{S}$ through rotation and translation, then discretize them into a volumetric representation using $c_{o}=[0,1,0]$. For both static and dynamic scenes, we discretize the vertices of each frame with $c_{s}=[0,0,1]$.

\noindent \textbf{Learning for UIV} To learn effective UIV features, we employ a pyramid-structured feature extractor \cite{lin2017feature} that captures multi-scale semantic information. Features from each scale are injected into corresponding downsampling stages of the network, enabling hierarchical fusion and improving the model’s capacity to represent complex interaction patterns.
\subsection{The UIV-Aligned Regularization}
\noindent \textbf{UIV-Aligned Motion Representation.} Our framework targets motion generation that is inherently aligned with the UIV. Traditional approaches \cite{tevet2022human, li2024controllable, xu2024regennet, jiang2024scaling} typically rely on forward kinematics, regressing joint rotations and constructing poses through kinematic chains. While effective for modeling articulated motion, these methods overlook the spatial context of the 3D environment, making it difficult to capture interactions with surrounding entities. To enable spatially grounded reasoning, we represent motion as voxel-based joint distributions over the UIV. Specifically, for each joint $k\in\{1,\cdots,K\}$ and time step $t$, we calculate a spatial probability map $\mathcal{P}_t^k$ over $\mathcal{S}$ to capture the likelihood of that joint occupying each voxel. This reformulation transforms motion generation into a structured spatial inference task, naturally aligned with the UIV representation. Following prior work on human pose estimation \cite{srivastav2024selfpose3d,liu2023mmda}, we model $\mathcal{P}_t^{k}$ as a normalized isotropic Gaussian distribution centered at the ground truth joint position $j^k_t\in\mathbb{R}^3$:
\begin{equation}
    \mathcal{P}_t^k(u)=\frac{1}{(2\pi\sigma^2)^{3/2}}\exp{(-\frac{\Vert u-j^k_t\Vert^2_2}{2\sigma^2})}
\end{equation}
where $\sigma>0$ denotes the standard deviation, controlling the spatial uncertainty and spread of the distribution. Given a predicted distribution $\hat{\mathcal{P}}^k_t$, we obtain the predicted joint position by computing the first moment (expectation) over the interactive volume:
\begin{equation}
\label{eq5}
    \hat{j}^k_t=\mathbb{E}_{\hat{\mathcal{P}}^k_t}[u]=\int_{u\in\mathcal{S}} \hat{\mathcal{P}}^k_t(u)\cdot u\,du
\end{equation}
This formulation is fully differentiable, making it easy to use in end-to-end learning. By spreading probability across nearby voxels, it avoids sparse gradients and helps the model learn even when predictions are not accurate. More importantly, computing the expectation over a continuous spatial distribution mitigates discretization errors. It allows sub-voxel precision in joint localization, enabling the model to produce fine-grained predictions that go beyond the native resolution of the voxel grid. This enhances both spatial accuracy and interaction fidelity within the UIV space. For visualization, we convert the predicted distributions into SMPL parameters.Joint coordinates are extracted via Eq.~\ref{eq5}, followed by SMPLify \cite{bogo2016keep} to estimate pose parameters.
The shape parameter is fixed to a default value.

\setlength{\abovecaptionskip}{2pt}  
\setlength{\belowcaptionskip}{0pt}
\begin{table*}[!t]
\centering
\caption{Quantitative comparison on the FullBodyManipulation dataset for the human-object interaction task.}
\begin{tabular}{cccccccccc}
\toprule
Methods & FS$\downarrow$ & FID$\downarrow$ & C-prec$\uparrow$ & C-rec$\uparrow$ & C-acc$\uparrow$ & C-F1$\uparrow$ & MPJPE$\downarrow$ & T-root$\downarrow$ \\\midrule
Interdiff \cite{xu2023interdiff} & 0.42 & 208.0 & 0.63 & 0.28 & - & 0.33 & 25.91 & 63.44 \\
MDM \cite{tevet2022human} & 0.48 & 6.16 & 0.72 & 0.47 & 0.62 & 0.53 & 17.86 & 34.16 \\
Lin-OMOMO \cite{li2023object} & 0.41 & 15.33 & 0.68 & 0.56 & - & 0.57 & 21.73 & 36.62 \\
Pred-OMOMO \cite{li2023object} & 0.40 & 4.19 & 0.73 & 0.66 & - & 0.66 & 18.66 & 28.39 \\
GT-OMOMO \cite{li2023object} & 0.41 & 5.69 & 0.77 & 0.66 & 0.74 & 0.67 & 15.82 & 24.75 \\
CHOIS \cite{li2024controllable} & \textbf{0.35} & 0.69 & 0.80 & 0.64 & 0.61 & 0.67 & 15.30 & 24.43\\
SemGeoMo \cite{cong2025semgeomo} & 0.57 & 1.17 & 0.84 & 0.74 & 0.85 & 0.77 & 16.62 & -\\
Ours & 0.39 & \textbf{0.51} & \textbf{0.91} & \textbf{0.86} & \textbf{0.93} & \textbf{0.86} & \textbf{12.15} & \textbf{11.20}\\\bottomrule
\end{tabular}
\label{table1}
\end{table*}

\begin{table*}[!t]
\centering
\caption{Quantitative comparison on the TRUMANS dataset for the human-scene interaction task.}
\begin{tabular}{ccccccc}
\toprule
Methods & Diversity$\rightarrow$ & FID$\downarrow$  & MPJPE$\downarrow$ & T-root$\downarrow$ & FS$\downarrow$ & Goal Dist.$\downarrow$\\\midrule
Real & 12.342 & - & - & - & 0.173 & - \\\midrule
Trumans \cite{jiang2024scaling} & 11.892 & 13.290 & 19.964 & 13.857 & 0.244 & 25.434 \\
Ours & \textbf{12.247} & \textbf{2.650} & \textbf{16.263} & \textbf{11.578} & \textbf{0.155} & \textbf{20.136}\\\bottomrule
\end{tabular}
\label{table3}
\end{table*}

\begin{table}[!t]
\centering
\caption{Quantitative comparison on the NTU120-AS dataset for the human-human interaction task.}
\resizebox{\columnwidth}{!}{
\begin{tabular}{cccc}
\toprule
Methods & FID$\downarrow$ & Diversity$\rightarrow$ & MultiModality$\uparrow$\\\midrule
Real & - & 23.265 & 17.695 \\\midrule
ReGenNet \cite{xu2024regennet} & 3.045 & 21.925 & \textbf{16.598}\\
Ours & \textbf{2.216} & \textbf{22.169} & 16.507\\\bottomrule
\end{tabular}}
\label{table2}
\end{table}
\noindent \textbf{Training Strategy.} Given the unified interactive volume introduced in Section~\ref{sec:unified}, it is natural to adopt a unified training strategy during model optimization. Specifically, we transform datasets involving human-human, human-object, and human-scene interactions into the UIV representation, and train the model on all three interaction scenarios across all datasets with a 1:1:1 mixing ratio by task type. This encourages consistent learning across different interaction types within a shared motion generation framework.
\noindent \textbf{Loss Function.} Due to the differentiability of the expectation operation in Equation.~\ref{eq5}, this formulation integrates smoothly into our diffusion-based generative framework. Specifically, apart from supervising the voxel-wise probability distribution, we impose an additional objective that directly supervises the predicted joint position $\hat{j}_t^k$, computed from the predicted distribution $\hat{\mathcal{P}}_t^k(u)$:
\begin{equation}
\mathcal{L}_{pos}(\hat{j}_t^k,j_t^k)=\Vert\hat{j}_t^k-j_t^k\Vert_2^2
\end{equation}
\begin{equation}
\mathcal{L}_{vel}(\hat{j}^k,j^k)=\sum_{t=1}^T\Vert\frac{d}{dt}(\hat{j}^k(t))-\frac{d}{dt}(j^k(t))\Vert_2^2
\end{equation}
Beyond supervising the predicted joint locations, we further enforce structural consistency by applying explicit constraints on the length and orientation of each skeleton:
\begin{equation}
\mathcal{L}_{sk}(\hat{\boldsymbol{s}}_t^n,\boldsymbol{s}_t^n)=\Vert\hat{\boldsymbol{s}}_t^n-\boldsymbol{s}_t^n\Vert_2^2
\end{equation}
where $\boldsymbol{s}_t^n=j_t^{\text{parent}(n)}-j_t^{\text{child}(n)}$ is the $n$-th skeleton. $j_t^{\text{parent}(n)}$ and $j_t^{\text{child}(n)}$ denote the positions of the parent and child joints connected by the $n$-th skeleton, respectively.

In many prior motion generation frameworks \cite{tevet2022human, xie2023omnicontrol}, the global orientation of the initial pose is normalized, aligning all motion sequences to a fixed forward-facing direction. While this simplifies the learning process, it limits the model’s ability to adapt motion to different environments, since all sequences start with the same heading. To address this, we avoid orientation normalization during training. Instead, we introduce an explicit supervision signal on the initial heading direction. This encourages the model to infer plausible orientations based on interactive cues from the surrounding scene, enabling more context-aware motion synthesis:
\begin{equation}
\mathcal{L}_{ori}(\hat{\boldsymbol{d}}_0,\boldsymbol{d}_0)=1-\frac{\hat{\boldsymbol{d}}_0\cdot\boldsymbol{d}_0}{\Vert\hat{\boldsymbol{d}}_0\Vert_2\Vert\boldsymbol{d}_0\Vert_2}
\end{equation}
Here, $\boldsymbol{d}_0$ denotes the orientation vector at the first frame. It is computed geometrically from the skeletal structure, using the vectors connecting the left and right hip joints to the root:
\begin{equation}
    \boldsymbol{d}_0 = \frac{\boldsymbol{s}_0^{lhip}}{\Vert\boldsymbol{s}_0^{lhip}\Vert_2}\times\frac{\boldsymbol{s}_0^{rhip}}{\Vert\boldsymbol{s}_0^{rhip}\Vert_2}
\end{equation}
Finally, we formulate the overall objective by aggregating all the aforementioned loss terms:
\begin{equation}
    \mathcal{L} = \mathcal{L}_{rec} + \lambda_1\cdot\mathcal{L}_{pos} + \lambda_2\cdot\mathcal{L}_{vel} + \lambda_3\cdot\mathcal{L}_{sk} + \lambda_4\cdot\mathcal{L}_{ori}
\end{equation}
where $\lambda_i$ $(i=1,2,3,4)$ is weighting coefficient that balances the corresponding loss terms.

\section{Experiment}
\noindent \textbf{Datasets.} We conduct experiments on three datasets corresponding to distinct interaction scenarios: {FullBodyManipulation} data-set \cite{ li2024controllable,li2023object}, which involves human–object interactions; {NTU120-AS} dataset \cite{xu2024regennet}, which captures human–human interactions; and {TRUMANS} dataset \cite{jiang2024scaling}, which focuses on human–scene interactions. The FullBodyManipulation dataset offers approximately 10 hours of synchronized motion data involving humans and 15 distinct objects. This dataset is employed both to train our interaction model and to assess the quality of the generated outcomes. NTU120-AS includes 8,118 multi-view sequences depicting human–human interactions across 26 action types. Following prior work \cite{xu2024regennet}, we adopt the cross-subject evaluation protocol using data from Camera 1, allocating half of the subjects for training and the rest for testing follow. The TRUMANS dataset comprises 15 hours of rich motion capture data with intricate human–scene interactions embedded in 3D environments featuring dynamic elements and densely arranged layouts. From this dataset, we select interaction segments annotated with textual descriptions, which are then divided into training, validation, and testing subsets in a 7:2:1 ratio.

\noindent \textbf{Evaluation Metrics.} We follow the evaluation protocol outlined by prior work \cite{li2024controllable,li2023object,xu2024regennet}. For the human interaction task, we evaluate the quality of the generated motion using the foot sliding score (FS) and the Fréchet Inception Distance (FID). To assess contact accuracy, we report precision (C-prec), recall (C-rec), and the F1 score (C-F1). In addition, to quantify the deviation from ground-truth motion, we report the mean per-joint position error (MPJPE) and the root joint translation error (T-root). For the human–human interaction task, we report the FID to assess motion quality, along with Diversity and MultiModality metrics to evaluate the variability of generated motions. For the human–scene interaction task, in addition to the metrics described above, we also adopt a distance-based measure (Goal Dist.) to evaluate how well the generated motions align with target positions. All distance-based metrics are reported in centimeters (cm).
\subsection{Implementation Details}
\begin{table}[t]
\centering
\caption{Ablation results for human-object interaction tasks on the FullBodyManipulation dataset. `w/o space dist' means directly regressing the kinematic parameters of motion.}
\resizebox{\columnwidth}{!}{
\begin{tabular}{cccccccc}
\toprule
Methods & FID$\downarrow$ & C-prec$\uparrow$ & C-rec$\uparrow$ & C-F1$\uparrow$ & MPJPE$\downarrow$ & T-root$\downarrow$ \\\midrule
w/o unified & 0.54 & 0.89 & 0.87 & 0.84 & 12.03 & 10.89\\
w/o space dist. & 179.50 & - & - & - & 33.30 & 355.53\\
Ours & 0.51 & 0.91 & 0.86 & 0.86 & 12.15 & 11.20\\\bottomrule
\end{tabular}}
\label{tab:ablation_manip}
\end{table}

\begin{table}[t]
\centering
\caption{Ablation results for human-scene interaction tasks on the TRUMANS dataset. `w/o space dist' means directly regressing the kinematic parameters of motion.}
\resizebox{\columnwidth}{!}{
\begin{tabular}{ccccccc}
\toprule
Methods & Diversity$\rightarrow$ & FID$\downarrow$  & MPJPE$\downarrow$ & T-root$\downarrow$ & FS$\downarrow$ & Goal Dist.$\downarrow$\\\midrule
Real & 12.342 & - & - & - & 0.173 & - \\\midrule
w/o unified & 12.782 & 3.517 & 16.493 & 11.237 & 0.204 & 19.659 \\
w/o space dist. & 5.017 & 66.003 & 34.311 & 12.925 & 0.448 & 23.414 \\
Ours & 12.247 & 2.650 & 16.263 & 11.578 & 0.155 & 20.136\\\bottomrule
\end{tabular}}
\label{tab:ablation_truman}
\end{table}

\begin{table}[t]
\centering
\caption{Ablation results for human-human interaction tasks on the NTU120-AS dataset. `w/o space dist' means directly regressing the kinematic parameters of motion.}
\begin{tabular}{cccc}
\toprule
Methods & FID$\downarrow$ & Diversity$\rightarrow$ & MultiModality$\uparrow$\\\midrule
Real & - & 23.265 & 17.695 \\\midrule
w/o unified & 2.247 & 22.530 & 16.866\\
w/o space dist. & 6170.243 & 23.075 & 22.027\\
Ours & 2.216 & 22.169 & 16.507\\\bottomrule
\end{tabular}
\label{tab:ablation_ntu}
\end{table}
\begin{figure*}[t]
  \centering
  \includegraphics[width=\linewidth]{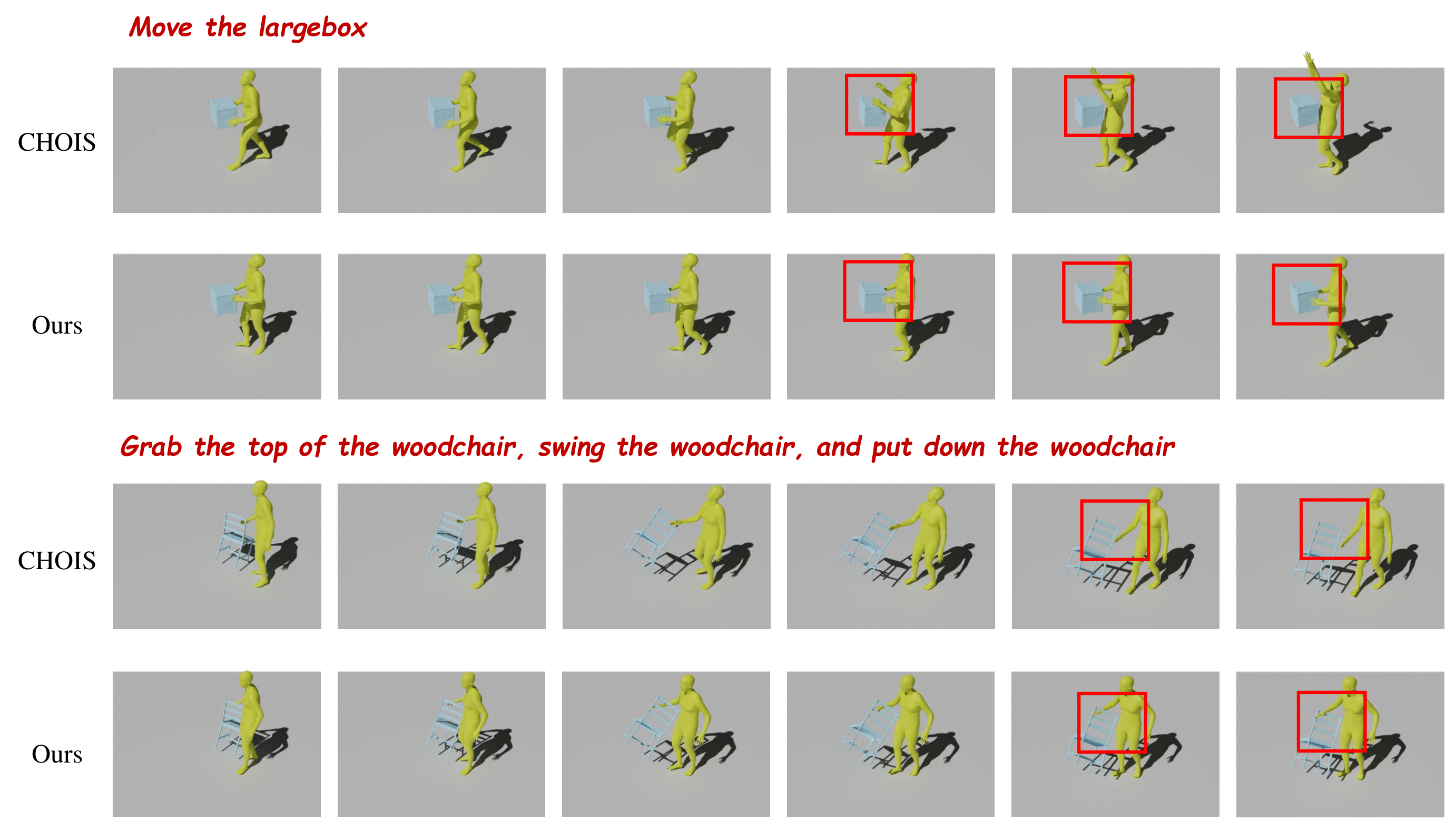}
  \caption{Qualitative comparison on the Human-Object Interaction dataset. Compared to state-of-the-art method CHOIS \cite{li2024controllable}, Uni-Inter achieves more precise control and interaction, particularly in hand movements. The blue object represents the conditional input, while the yellow-green person shows the generated motion.}
  \label{fig:manip}
\end{figure*}
\begin{figure*}[t]
  \centering
  \includegraphics[width=0.95\linewidth]{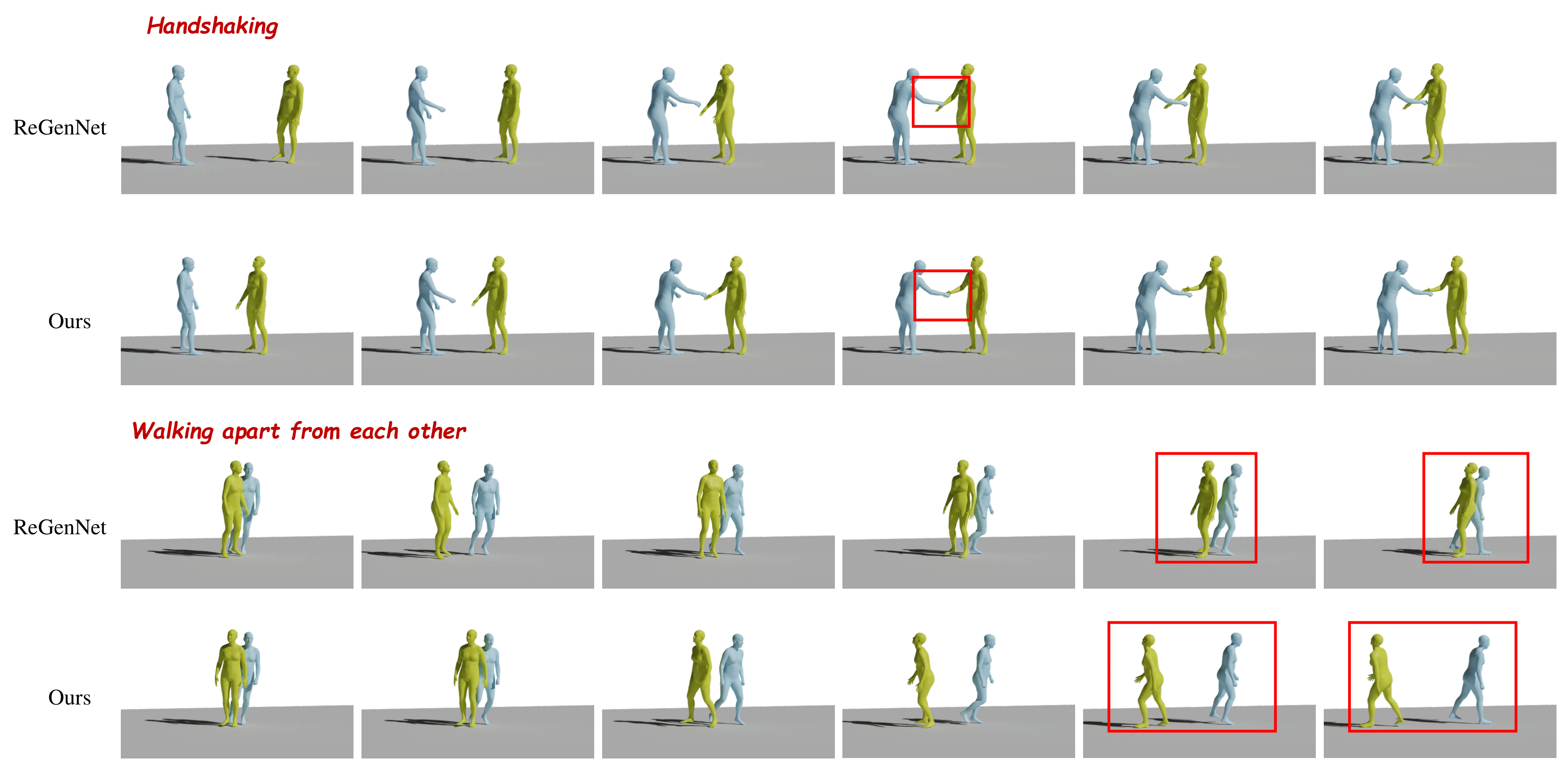}
  \caption{Qualitative comparison on the Human-Human Interaction dataset. Compared to ReGenNet \cite{xu2024regennet}, Uni-Inter demonstrates better spatial alignment of interaction events, resulting in more realistic and context-consistent motion generation. The blue person represents the conditional input, while the yellow-green person shows the generated motion.}
  \label{fig:ntu}
\end{figure*}
\noindent \textbf{Model Architecture.} We adopt a diffusion-based architecture for motion generation, where, as described in Section.~\ref{diffusion}, the model directly predicts $x_0$. The motion is modeled as Gaussian distribution maps, with the final output denoted as $x_0 = P$. Our model takes both text and UIV as conditional inputs. The textual input is first processed by a CLIP \cite{radford2021learning} encoder to extract text features $\mathcal{F}_{text}\in\mathbb{R}^{512}$ before being fed into the model. Further architectural details are provided in the supplementary material.
\begin{equation}
    P=\{\mathcal{P}_t^k\}_{k,t}\in\mathbb{R}^{T\times H\times W\times D\times K}
\end{equation}
The motion sequence $m_{1:T}$ can be obtained by directly taking the expectation over the output distribution $P$ as defined in Equation.~\ref{eq5}. For all experiments, we set the sequence length to $T=40$ frames.

\noindent \textbf{Training Details.} During training, all datasets from the three tasks are combined with equal weighting (1:1:1 ratio). In our experiments, we set $\sigma=3$ and $H=W=D=48$, corresponding to a physical space of 2.4 meters in height and 4.8 meters in both depth and width. Details of the ablation study on $\sigma$ are provided in the supplementary materials. The diffusion process is configured with 1,000 timesteps, and we adopt the DDIM \cite{song2020denoising} sampling strategy during inference. The model is trained using a total loss $\mathcal{L}$, with the weighting hyperparameter set to $\lambda_1=\lambda_2=\lambda_3=0.1$, $\lambda_4=1$. We use a learning rate of $3\times10^{-5}$, which decays linearly over training iterations. The model is trained for a maximum of 500,000 steps with a batch size of 32.
\subsection{Quantitative Results}
We evaluate our model on three tasks, benchmarking it against SOTA methods on each dataset.

\noindent \textbf{Human-Object Interaction.} Since Uni-Inter conditions on the object pose, it does not explicitly generate object pose sequences. Instead, our evaluation primarily focuses on the quality of human interaction motion generation. As shown in Table.~\ref{table1}, our method outperforms SOTA approaches on most evaluation metrics. By representing the physical environment as UIV, our model explicitly captures spatial layout and object configurations, enabling it to generate more realistic contact interactions, as evidenced by high contact precision (C-prec: 0.91), recall (C-rec: 0.86), and F1 score (C-F1: 0.86). In addition, a low FID of 0.51 indicates superior overall motion quality. Compared to CHOIS \cite{li2024controllable}, our model exhibits lower error, with a 54\% reduction in motion trajectory error, highlighting its effectiveness in capturing accurate dynamics.
\noindent \textbf{Human-Human Interaction.} For the human–human interaction task, we compare Uni-Inter against ReGenNet \cite{xu2024regennet}. Since only the evaluation code and pretrained weights for motion feature extraction are publicly available, we report results on FID, Diversity, and MultiModality. As shown in Table.~\ref{table2}, our model achieves a lower FID score, indicating higher motion quality, and demonstrates improved Diversity compared to the baseline. We further evaluate on the Chi3D-AS dataset \cite{xu2024regennet}, with detailed results provided in the supplementary material.
\noindent \textbf{Human-Scene Interaction.} We evaluate the human–scene interaction task on the TRUMANS \cite{jiang2024scaling} dataset, as shown in Table.~\ref{table3}. Our method is compared against state-of-the-art approaches and achieves superior performance across all metrics. Notably, our model generates higher-quality motions with an FID of 2.650 and a foot sliding score of 0.155. Furthermore, the distance between generated motions and target positions is improved by 20.8\% relative to the baseline. This improvement is largely attributed to the unified representation, which provides an explicit spatial structure that enhances the model’s understanding of scene geometry.

\begin{figure*}[t]
  \centering
  \includegraphics[width=0.95\linewidth]{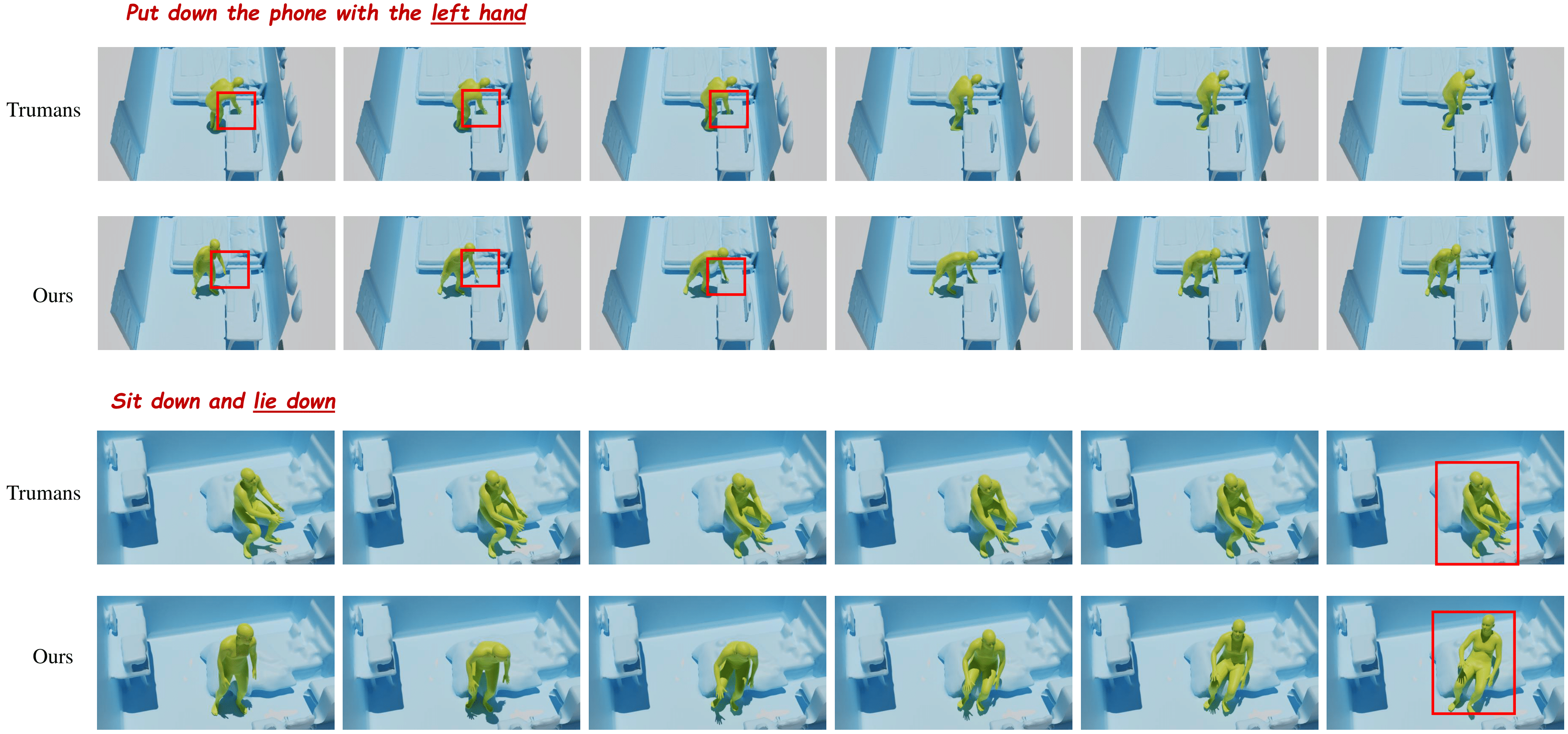}
  \caption{Qualitative comparison on the Human-Scene Interaction dataset. Compared to the SOTA method Trumans, Uni-Inter shows superior semantic understanding. In the first example, the key instruction is “left hand,” but Trumans incorrectly uses the right hand. In the second example, the key verb is “lie down,” which Trumans fails to execute, highlighting Uni-Inter’s advantage in accurately following semantic cues.}
  \label{fig:trumans}
\end{figure*}
\begin{figure*}[!t]
  \centering
  \includegraphics[width=0.9\linewidth]{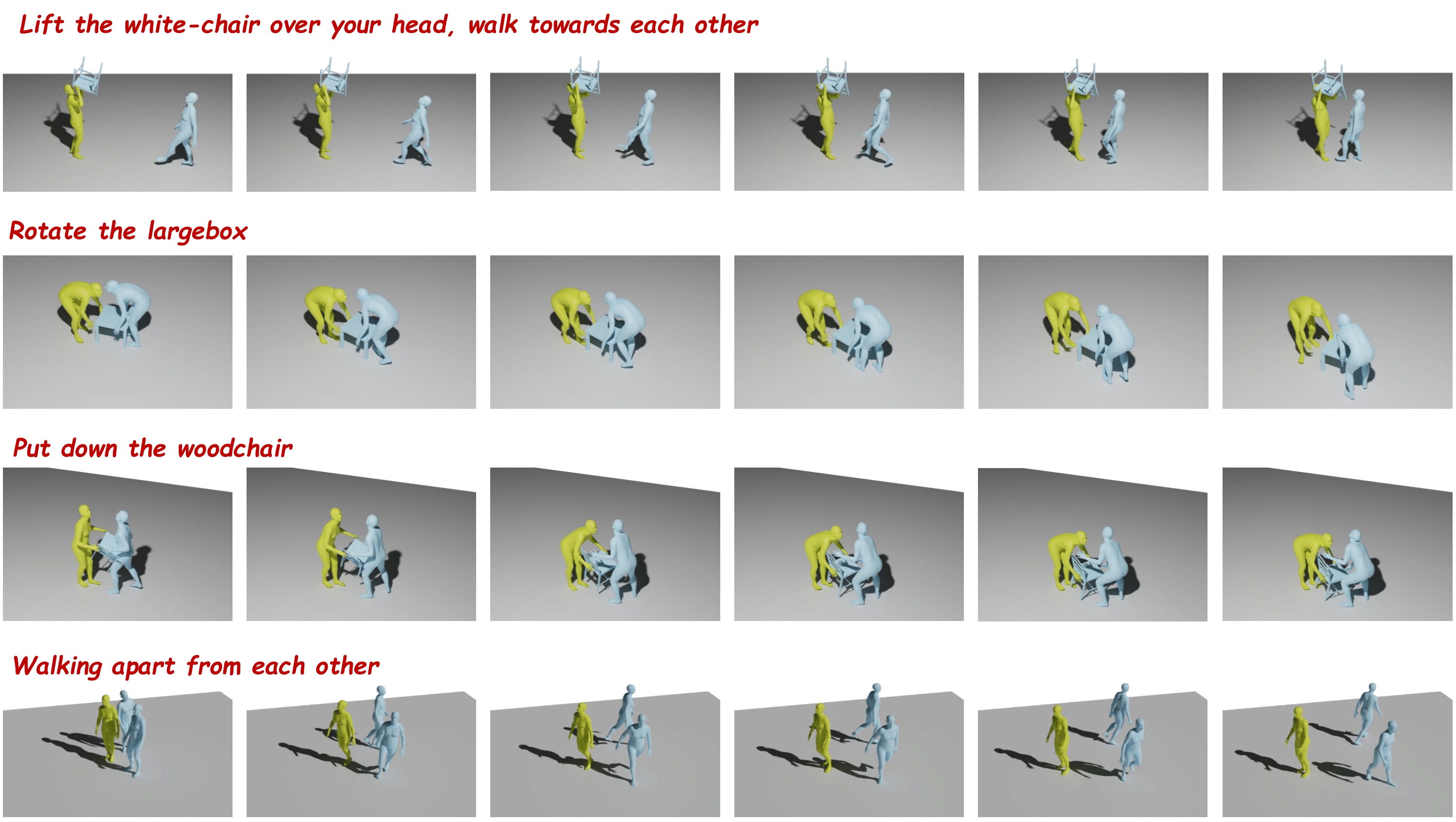}
  \caption{Qualitative results of compound interaction generation by Uni-Inter. Thanks to the unified modeling of humans, objects, and scenes via UIV, Uni-Inter supports high-quality compound motion generation across arbitrary combinations of interactive entities. Blue person and object indicate the input conditions, while yellow-green motions represent the generated results.}
  \label{fig:unified}
\end{figure*}

\subsection{Qualitative Results}
Our proposed method demonstrates substantial improvements over current state-of-the-art approaches, as evidenced by qualitative visualizations. These results clearly show that our method generates more natural interactions and achieves better text-motion alignment. Such performance gains can be attributed to the unified training framework and the introduction of UIV modeling, which significantly improves the accuracy of interaction modeling. We provide qualitative comparisons across three benchmark tasks. As shown in Figure.~\ref{fig:manip}, for human-object interaction, our method outperforms the SOTA method CHOIS \cite{li2024controllable} in terms of text-conditioned motion control, particularly excelling in the fine-grained synthesis of hand movements. Figure.~\ref{fig:ntu} illustrates human-human interaction results, where our approach exhibits superior coordination and consistency compared to ReGenNet \cite{xu2024regennet}. For human-scene interaction, Figure.~\ref{fig:trumans} presents a comparison with the Trumans \cite{jiang2024scaling} method, demonstrating that Uni-Inter offers better semantic understanding and motion alignment with the input text. In addition, Uni-Inter handles diverse interactive entities in a unified manner, enabling simultaneous interactions with multiple entities, as illustrated in Figure~\ref{fig:unified}. We showcase compound interaction examples that highlight Uni-Inter’s ability to flexibly handle diverse entity combinations within a single, unified motion generation process. More visualizations are provided in the supplementary material. 
\subsection{Ablation Studies}
\begin{table}[t]
\centering
\caption{Ablation study on loss terms for human-object interaction on the FullBodyManipulation dataset.}
\resizebox{\columnwidth}{!}{
\begin{tabular}{cccccccc}
\toprule
Methods & FID$\downarrow$ & C-prec$\uparrow$ & C-rec$\uparrow$ & C-F1$\uparrow$ & MPJPE$\downarrow$ & T-root$\downarrow$ \\\midrule
w/o $\mathcal{L}_{pos}$ & 0.55 & 0.91 & 0.79 & 0.80 & 12.01 & 11.59\\
w/o $\mathcal{L}_{vel}$ & 0.58 & 0.85 & 0.66 & 0.67 & 12.82 & 12.11\\
w/o $\mathcal{L}_{sk}$ & 0.67 & 0.88 & 0.67 & 0.68 & 12.30 & 12.80\\
w/o $\mathcal{L}_{ori}$ & 0.67 & 0.91 & 0.82 & 0.83 & 12.21 & 11.58\\
Ours & 0.51 & 0.91 & 0.86 & 0.86 & 12.15 & 11.20\\\bottomrule
\end{tabular}}
\label{tab:loss_manip}
\end{table}

\begin{table}[t]
\centering
\caption{Ablation study on loss terms for human-scene interaction on the TRUMANS dataset.}
\resizebox{\columnwidth}{!}{
\begin{tabular}{ccccccc}
\toprule
Methods & Diversity$\rightarrow$ & FID$\downarrow$  & MPJPE$\downarrow$ & T-root$\downarrow$ & FS$\downarrow$ & Goal Dist.$\downarrow$\\\midrule
Real & 12.342 & - & - & - & 0.173 & - \\\midrule
w/o $\mathcal{L}_{pos}$ & 12.229 & 3.188 & 16.577 & 11.543 & 0.145 & 19.742 \\
w/o $\mathcal{L}_{vel}$ & 13.158 & 2.708 & 16.784 & 12.218 & 0.176 & 20.334 \\
w/o $\mathcal{L}_{sk}$ & 12.641 & 4.016 & 16.808 & 12.585 & 0.153 & 20.852 \\
w/o $\mathcal{L}_{ori}$ & 12.474 & 3.108 & 16.908 & 11.625 & 0.171 & 20.284 \\
Ours & 12.247 & 2.650 & 16.263 & 11.578 & 0.155 & 20.136\\\bottomrule
\end{tabular}}
\label{tab:loss_truman}
\end{table}

\begin{table}[t]
\centering
\caption{Ablation study on loss terms for human-human interaction on the NTU120-AS dataset.}
\begin{tabular}{cccc}
\toprule
Methods & FID$\downarrow$ & Diversity$\rightarrow$ & MultiModality$\uparrow$\\\midrule
Real & - & 23.265 & 17.695 \\\midrule
w/o $\mathcal{L}_{pos}$ & 7.582 & 21.609 & 16.516\\
w/o $\mathcal{L}_{vel}$ & 5.811 & 22.038 & 16.824\\
w/o $\mathcal{L}_{sk}$ & 11.413 & 21.799 & 16.512\\
w/o $\mathcal{L}_{ori}$ & 4.214 & 21.593 & 16.673\\
Ours & 2.216 & 22.169 & 16.507\\\bottomrule
\end{tabular}
\label{tab:loss_ntu}
\end{table}

\noindent \textbf{Ablation on Unified Training.} To assess the effectiveness of the proposed unified representation, we conduct ablation studies, with results summarized in Table.~\ref{tab:ablation_manip}, Table.~\ref{tab:ablation_truman} and Table.~\ref{tab:ablation_ntu}. Note that the `w/o unified' setting is trained on a single specific dataset without scenario integration. Across all three interaction tasks, the unified training scheme consistently improves motion quality, as reflected by lower FID and foot sliding scores. For the human-object interaction task, the model achieves a higher contact F1 score (C-F1 = 0.86 vs. 0.84), indicating improved interaction fidelity. In the human-scene interaction task, both FS and FID show significant gains, suggesting that multi-task training provides strong auxiliary benefits by enhancing the model’s reasoning capabilities.

\noindent \textbf{Ablation on Motion Representation.} Unlike previous approaches \cite{tevet2022human, li2024controllable, xu2024regennet, jiang2024scaling} that directly predict the kinematic parameters of motion, our model outputs a spatial probability distribution, enabling it to reason over spatial locations and explicitly capture the underlying correlation between motion and spatial context. To evaluate the necessity of this design, we perform an ablation study by replacing the distributional output with direct kinematic motion sequence regression. As shown in Table.~\ref{tab:ablation_manip}, Table.~\ref{tab:ablation_truman} and Table.~\ref{tab:ablation_ntu}, this modification leads to a substantial performance drop, particularly evident in the FID metric. These results underscore the critical role and irreplaceability of our proposed spatial distribution formulation.

\noindent \textbf{Ablation on Loss Function.} We conduct a systematic evaluation of our model's loss design in Table.~\ref{tab:loss_manip}, Table.~\ref{tab:loss_truman} and Table.~\ref{tab:loss_ntu}. For the human-object interaction task, removing either the skeleton consistency loss $\mathcal{L}_{sk}$ or the velocity loss $\mathcal{L}_{vel}$ significantly degrades interaction quality, as indicated by a drop in the contact F1 score from 0.86 to 0.68 and 0.67, respectively. Furthermore, the orientation loss $\mathcal{L}_{ori}$ is shown to be crucial for maintaining overall motion fidelity, with FID increasing from 0.51 to 0.67 when it is removed. Similarly, for both the human-human and human-scene interaction tasks, $\mathcal{L}_{sk}$ remains the most significant component. Its removal results in a substantial increase in FID, highlighting its necessity in guiding the generation of high-quality motions.

\section{Conclusion}
In this work, we presented Uni-Inter, a unified framework for human motion generation across human-human, human-object, and human-scene interactions. By leveraging a unified interactive volume representation, Uni-Inter abstracts heterogeneous entities into a shared interactive volume, enabling consistent reasoning and motion synthesis across diverse and compound interaction scenarios. This unified design enables strong generalization to complex multi-entity interactions, while maintaining scalability and modularity. UIV-aligned Regularization is introduced to guide the model in better learning the semantic structure and spatial relationships encoded in the UIV. Extensive experiments across three representative interaction domains demonstrate that Uni-Inter delivers coherent, semantically aligned motions and outperforms existing task-specific baselines both qualitatively and quantitatively. In future work, we aim to extend Uni-Inter to a causal, real-time setting where motion is generated based on partial observations, supporting dynamic planning in interactive environments. This direction opens new possibilities for deploying Uni-Inter in interactive graphics, embodied agents, and real-time virtual experiences.
\bibliographystyle{ACM-Reference-Format}
\bibliography{main}